**Bennett, CC, Doub TW, and R Selove. (2012). EHRs Connect Research and Practice: Where Predictive Modeling, Artificial Intelligence, and Clinical Decision Support Intersect.** *Health Policy and Technology*. **In Press.**
http://www.sciencedirect.com.ezproxy.lib.indiana.edu/science/article/pii/S221188371200038X

# EHRs Connect Research and Practice: Where Predictive Modeling, Artificial Intelligence, and Clinical Decision Support Intersect

Casey C. Bennett, M.A.[1,2], Thomas W. Doub, Ph.D.[1], Rebecca Selove, Ph.D.[1]

[1] Centerstone Research Institute
Nashville, TN

[2] School of Informatics and Computing
Indiana University
Bloomington, IN

**Author Contact Info:**
Casey Bennett
Lead Data Architect
Dept. of Informatics
Centerstone Research Institute

365 South Park Ridge Road
Suite 103
Bloomington, IN 47401
(812)337-2302
Casey.Bennett@CenterstoneResearch.org

Thomas Doub
Chief Operating Officer (COO)
Centerstone Research Institute

44 Vantage Way
Suite 280
Nashville, TN 37208
(615)460-6600
Tom.Doub@CenterstoneResearch.org

Rebecca Selove
Scientific Grant Writer
Centerstone Research Institute

44 Vantage Way
Suite 280
Nashville, TN 37208
(615)460-6600
RebeccaSelove@CenterstoneResearch.org

**EHRs Connect Research and Practice: Where Predictive Modeling, Artificial Intelligence, and Clinical Decision Support Intersect**

**Research highlights**

- EHRs are increasingly likely to contain data and functionality that can support computational approaches to healthcare.

- Predictive modeling of EHR data has achieved 70-72% accuracy in predicting individualized treatment response at baseline.

- Clinical decision support can be conceptualized as a form of artificial intelligence embedded within clinical systems.

- Despite challenges, data-driven clinical decision support based on real-world populations offers numerous advantages.

- Such approaches may also contribute to better implementation of research into real-world clinical practice.

**Abstract**

**Objectives:** Electronic health records (EHRs) are only a first step in capturing and utilizing health-related data – the challenge is turning that data into useful information.  Furthermore, EHRs are increasingly likely to include data relating to patient outcomes, functionality such as clinical decision support, and genetic information as well, and, as such, can be seen as repositories of increasingly valuable information about patients' health conditions and responses to treatment over time.   **Methods:** We describe a case study of 423 patients treated by Centerstone within Tennessee and Indiana in which we utilized electronic health record data to generate predictive algorithms of individual patient treatment response.    Multiple models were constructed using predictor variables derived from clinical, financial and geographic data.  **Results:**  For the 423 patients, 101 deteriorated, 223 improved and in 99 there was no change in clinical condition.    Based on modeling of various clinical indicators at baseline, the highest accuracy in predicting individual patient response ranged from 70-72% within the models tested.   In terms of individual predictors, the Centerstone Assessment of Recovery Level – Adult (CARLA) baseline score was most significant in predicting outcome over time (odds ratio 4.1 + 2.27).  Other variables with consistently significant impact on outcome included payer, diagnostic category, location and provision of case management services. **Conclusions:** This approach represents a promising avenue toward reducing the current gap between research and practice across healthcare, developing data-driven clinical decision support based on real-world populations, and serving as a component of embedded clinical artificial intelligences that "learn" over time.

**Introduction**

      In the modern Information Age, we are often overwhelmed by data.  The challenge is converting that data into useful information.[1]  Practitioners and organizations in the healthcare field face this challenge as they strive to translate new technological and research advances into clinical advances.  Collecting data in an electronic health record (EHR) is only the first step – indeed, we must further leverage that data through technology in order to provide useable, actionable information.  Without such additional functionality, EHRs are essentially paper-based records stored in electronic form, and their potential to transform care is limited.  Expanding use of EHRs by modeling and transforming that data has broad implications for connecting research and practice in the future.

      The lack of actionable, predictive information in healthcare is ironic given that it is a particularly data-rich environment. Healthcare is rife with documentation requirements established by payers and accreditation bodies that produce much potentially useful information. Managed care companies and disease management vendors have large teams of data analysts sifting through health-related claims data (often limited to items such as service date, service type, and diagnosis) in order to identify people who may be appropriate for outreach and intervention to prevent negative outcomes.  In contrast, many health providers, who produce much richer datasets that include information about symptoms, clinical assessments, patient behaviors, and social factors such as SES (socio-economic status), level of education, and employment, lack the capacity to analyze and transform such data into actionable information. This represents missed opportunities for significant data-driven healthcare improvements based on the effective utilization of data in EHRs.  In this paper, we provide an argument and real-world example of what this might entail.

      Data-driven healthcare is key for addressing the known 13-17 year gap between research and practice in clinical service delivery.[2]  Connecting the processes of collecting data in the course of providing services to ongoing modeling efforts could address two problems: evidence-based guidelines derived from research are often out-of-date by the time they reach widespread use;  and such guidelines don't always account for real-world variations and comorbidities that can impede effective implementation.[3,4] While clinicians are receptive to meaningful advances in treatment, they are also aware of the complexities they face in daily practice that often are not effectively addressed by evidence-based guidelines. For example, guidelines may recommend the same treatment for everyone (i.e., one-size-fits-all), or prescribe a rigorous, standardized sequence of treatment options.[3,4] Standardized algorithmic approaches to care are clearly effective in research settings; however, clinicians frequently lack the time and

information required to make a reliable diagnosis (a prerequisite for many algorithms) and to deal with the minutiae of specific treatment recommendations.[5]

Concerns that treatment algorithms are inflexible and unresponsive to the needs of real-world clinical populations appear to have a basis in fact. A strong allegiance to high fidelity often overshadows the local adaptation that often must occur in live clinical practice. Indeed, even when implemented using technological methods, static, one-size-fits-all algorithms do not take full advantage of the technology, or the potential value of the live clinical data surrounding them in EHRs. Such algorithms are essentially "ignorant" of the real-world knowledge sitting right next to them – knowledge that could be used to adapt them to their "environment". From an evolutionary perspective, the end result would likely be extinction. Not surprisingly, when artificial implementation supports are removed, practices often fail to sustain or deviate significantly from fidelity.[6]

The use of predictive models for informing healthcare treatment algorithms accentuates the tension that exists between the art and science of treating common health disorders, between the knowledge of experienced clinicians and predictive recommendations derived from data. This old controversy is best characterized by Paul Meehl, who noted nearly half a century ago "When you are pushing [scores of] investigations [140 in 1991], predicting everything from the outcomes of football games to the diagnosis of liver disease and when you can hardly come up with a half dozen studies showing even a weak tendency in favor of the clinician, it is time to draw a practical conclusion" (p.372-373).[7]

The research-practice gap clearly contributes to the push for clinical decision support in healthcare. The simple concept underlying this effort is that providing decision support to clinicians will improve their decision making, leading to better efficiency and quality in care.[8-10] Decision support, as the name implies, refers to providing information to clinicians, typically at the point of decision making. It comes in a variety of forms[11] and has also been applied to problems related to production, quality, and infrastructure across many fields.[12-14] However, many current decision support systems in healthcare rely on expert- or standards-based models, rather than models that adapt population-based guidelines to individual patient characteristics by utilizing existent EHR patient data. The former are based on statistical averages or expert opinion of what works for groups of people in general, whereas data-driven models are essentially an individualized form of practice-based evidence drawn from the live population. The latter falls within the concept of "personalized medicine."

There are numerous advantages to building decision support systems driven by live data based on the actual population. The ability to adapt specific treatments to fit the specific symptoms and functional characteristics of an individual's disorder transcends the traditional disease model. Much of the focus of the past decade has been on the utility of genetic data to inform individualized care (a.k.a "personalized medicine").[15,16] However, it is likely that the next decade will focus on the use of multiple sources of data – genetic, clinical, socio-demographic – to build a more complete profile of the individual, their inherited risks, and the environmental/behavioral factors associated with disorder and the effective treatment thereof.[17] Indeed, this is already apparent in the trend of combining clinical and genetic indicators for predicting cancer prognosis.[18-19] It should be expected that much of this data will be derived from 21st century EHRs.

Improving quality and use of clinical decision support tools requires accurately anticipating the consequences of various choices providers make during clinical service delivery. This entails understanding what indicators are important *in general*, as well as the nuances of what is important in *a particular set of circumstances* (e.g., the impact of alternative treatments on an individual patient's outcome trajectory). Predictive modeling and data mining (PM/DM) are two interrelated approaches to this issue. Data mining is the process of discovering patterns in data, typically through automatic or semi-automatic means (also sometimes referred to as machine learning).[20] It can be applied to regression and classification problems, identify linear and nonlinear patterns, and adapted to binary or multi-class outcomes. Identified patterns can then be used to make predictions about future events, i.e., predictive modeling. We may want to know what treatment will result in the best outcome, or how staff should be scheduled for maximum efficiency. Data mining methods can answer both of these questions. The end result is *actionable* information for clinicians and managers.

For instance, we can use historical patients in the existent EHR to identify patterns that can then be applied to new clients as they walk in the door (e.g., patterns that typify patients who responded to a particular treatment and those who did not). The PM/DM approach utilizes sophisticated techniques to "decompose" individuals into their component characteristics and make targeted predictions based on the probabilities associated with those components (in essence creating a composite score from individual component probabilities) – even for individuals with novel sets of characteristics – rather than basing predictions on statistical averages or case matching. The measure of the ability to successfully accomplish this is a pragmatic one – the ability to predict accurately on sets of instances outside the training sample used to construct the model.[20]

However, data mining or clinical decision support alone is not sufficient for delivering personalized recommendations at the point of care.  Utilizing them in conjunction can create a system of real-time data-driven clinical decision support, or "adaptive" clinical decision support.   The result is a more responsive and personalized model. One advantage of this approach is that the system can improve over time by evaluating its predictions and "learning" from its mistakes.  In a sense, this represents a form of artificial intelligence embedded within the live clinical system.  An adaptive model can generate and apply new evidence about effective practices every day within a live clinical system (i.e., "practice-based evidence"), as well as modify clinically-established guidelines to fit the needs of patients in real world settings.

Continuous improvement of clinical decision support and advancement of clinical knowledge are seen as key features for future data systems and EHRs in healthcare.[9] In terms of actual application, modeling can be used to support clinical decisions provided a flexible, adaptable IT framework can consolidate data from different sources. Typically, data warehousing provides such an infrastructure to mix rich genetic and clinical data, mine the data, and develop algorithms.  In contrast to the EHR, a data warehouse does not have to be tied to a single provider organization, and thus enhances the power, scope, and utility of the underlying EHRs.  In many cases, Health Information Exchanges (HIEs) can be designed as data warehouses.  Predictive algorithms can be derived from these data warehouses as well as the individual underlying EHRs (Fig.1 Time 1) and then be applied to data obtained from new patients, with the end result pushed to the front-end web application of the EHR and displayed for clinicians' use along with the clinical record (Fig.1 Time 2).  Patterns learned from past patients' experiences can be continuously refined as new patients enter the system.

**Figure 1: Clinical Decision Support - Data Flow Diagram**

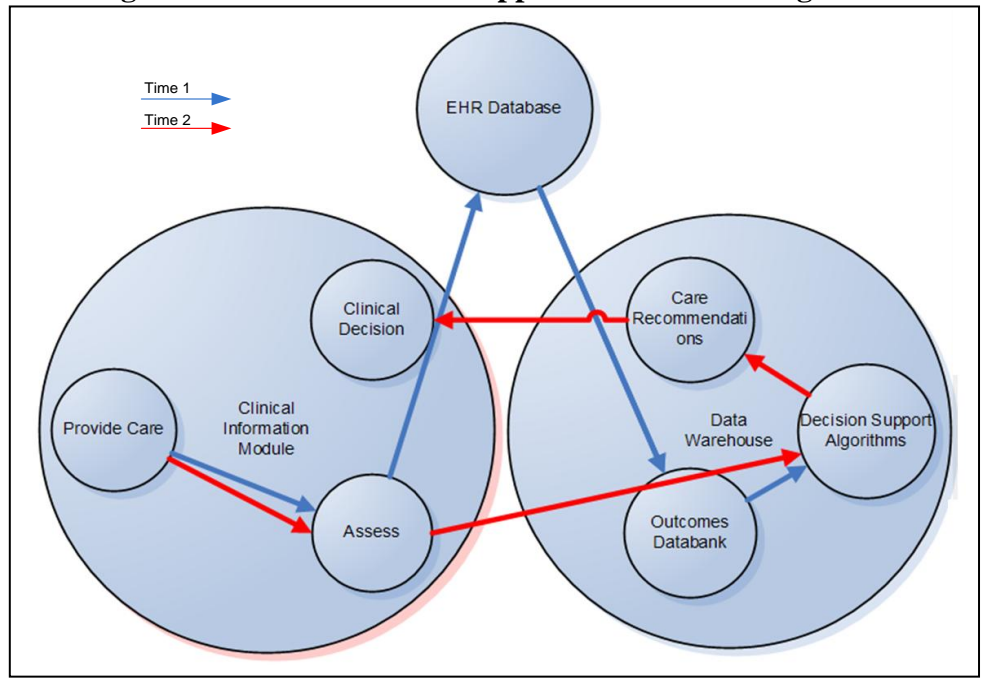

Real-time data-driven clinical decision support narrows the research-practice gap and incorporates the idiosyncrasies of the actual clinical population, melding research and practice into the same activity by utilizing clinical EHR data in an ongoing basis to create and adapt knowledge. One is no longer limited to do research, only later to try to apply it to practice. Research *becomes* practice.

The following case study demonstrates the challenges and opportunities of utilizing these aforementioned methods on EHR data to develop optimal treatment recommendations in a real-world clinical population.

**Methods**

**A. Setting**

Centerstone providers in Tennessee and Indiana see over 70,000 distinct patients a year across over 130 outpatient clinical sites. Centerstone Research Institute (CRI) is an arm of Centerstone devoted to integrating evidence and practice, conducting clinical research, developing clinical decision support tools, and building new healthcare informatics technologies, among other goals. Centerstone, which has a fully functional EHR that maintains virtually all relevant patient records, operates under a mixture of fee-for-service and case rate payment methodologies, including Medicare, Medicaid, and commercial payers, as well various other payers such as county subsidies, DCS (Tennessee Dept. of Children Services), federal probation funds, and grants. Centerstone, like many other community mental health providers, is under increasing pressure from all payer sources to hold down costs and increase service provision.

The initial work described here was necessitated by changes to a state-run payer (non-Medicaid "Safety Net") in Tennessee, which compelled Centerstone to better optimize the match between available services and the clinical needs of patients so as to minimize provision of unnecessary services while maximizing outcomes. We approached this requirement by determining the probability that a given set of services would result in average or above-average outcomes for a particular patient. This allowed us to identify services that would provide patients with the best probability of positive outcomes while minimizing use of services unlikely to be beneficial, increasing the availability of limited resources for other patients.

**B. Data**

Data was extracted from Centerstone's electronic health record into a specialized schema in the data warehouse for data mining applications. The target variable was the follow-up CARLA outcome measure (Centerstone Assessment of Recovery Level – Adult, http://centerstoneresearch.org/files/CARLA_instrument.pdf) at 6 months post baseline. The CARLA is a measure of level of recovery developed and validated by clinical experts at Centerstone, informed by recovery levels used by Pike's Peak Mental Health Center, as well as other level-of-care models including ASAM, LOCUS, and Ohio scale.[21] Using the CARLA, clinicians provide a systematic rating of patient symptoms, functioning, supports, insight, and engagement in treatment - a score of 1 indicating severe impairment in each dimension and a score of 4 indicating little or no impairment. Descriptive statistics on CARLA at baseline, final (follow-up), and delta (change in health outcome over time) can be seen in Table 1. Table 2 shows frequencies for different categories of change (deterioration, improvement), with roughly 54% showing improvement averse to 24% showing deterioration.

## Table 1: Carla Descriptive Stats

|  | N | Minimum | Maximum | Mean | Std. Deviation |
|---|---|---|---|---|---|
| Baseline CARLA | 423 | 1.2 | 3.6 | 2.428 | .4069 |
| Final CARLA | 423 | 1.4 | 3.8 | 2.603 | .3955 |
| CARLA Delta | 423 | -1.0 | 1.8 | .175 | .4190 |

**Table 2: CARLA Delta Frequencies**

|  | Frequency | Percent | Valid Percent | Cumulative Percent |
|---|---|---|---|---|
| Deteriorated | 101 | 23.9 | 23.9 | 23.9 |
| No Change | 99 | 23.4 | 23.4 | 47.3 |
| Improved | 223 | 52.7 | 52.7 | 100.0 |
| Total | 423 | 100.0 | 100.0 |  |

Predictor variables initially incorporated into the predictive model included Baseline CARLA Score, Gender, Race, Age, Baseline Tennessee Outcomes Measurement System (TOMS) Symptomatology Score, Baseline TOMS Functioning Score, Previous Mobile Crisis Encounter (binary, yes/no), Diagnosis Category, Payer, Location, County, Region Type (Urban or Rural), Service Profile (types of services received) and Service Volume (amount of services received). The initial sample was delimited to June 1, 2008 through approximately June 1, 2009 and included only new intakes at time of baseline CARLA (had not seen previously in Centerstone's clinics since at least 2001). After these various filters were applied and data was screened for missing key fields (such as the CARLA at both baseline and follow-up), the final sample size for initial modeling was 423. Basic demographics of this sample can be seen in Table 3 – the patients were largely white, female, and suffering from mood disorders, with a significant number (62%) exhibiting co-occurring chronic physical disorders such as chronic pain, high blood pressure, diabetes, and cardiovascular disease.

**Table 3: Patient Demographics**

| Gender | % | Race | % | Age Group | % | Diagnosis | % | Locale | % |
|---|---|---|---|---|---|---|---|---|---|
| F | 71.3% | White | 79.5% | 18-29 | 27.4% | Anxiety | 8.0% | Urban | 44.4% |
| M | 28.7% | Black | 10.1% | 30-45 | 37.7% | Bipolar | 20.5% | Rural | 55.6% |
|  |  | Asian | 0.9% | 46-65 | 31.0% | Depression | 42.8% |  |  |
|  |  | Other | 1.8% | >65 | 3.9% | Other | 28.7% |  |  |
|  |  | Unknown | 7.6% |  |  |  |  |  |  |

**C. Data Mining**

After the initial data loading and calculations were made, data was loaded into KNIME (Version 2.1.1)[22], an advanced data mining, modeling, and statistical software. Data mining typically follows a standard process flow that can be broken into a number of main steps: data preparation, feature selection, model construction, and model evaluation. It should be noted that not all steps are performed every time – for instance one may build models without any feature selection in order to evaluate the effect of feature selection on a particular dataset. Below, these steps are briefly outlined in the context of the current study; a more comprehensive overview of specific data mining strategies and methodology can be found in other resources on the subject.[20,23]

The first stage is data preparation. The initial analysis focused on clinical outcomes as measured by the change in CARLA scores over time. The primary question was whether patients would obtain average, better, or worse outcomes based on services received. As such, the target variable was discretized into a binary variable of plus/minus the mean (equivalent to an equal bins classification approach). The consequences and assumptions of reduction to a binary classification problem are addressed in Boulesteix et al.,[19] who noted that the issues of making such assumptions are roughly equivalent to those around normal distributions. All predictor variables were z-score normalized. Subsequently, all predictor variables were either 1) not discretized (labeled "Bin Target"), or 2) discretized via CAIM (Class-Attribute Interdependence Maximization). CAIM is a form of entropy-based discretization that attempts to maximize the available "information" in the dataset by delineating categories in the predictor variables that relate to classes of the target variable. Use of patterns in the data itself has been shown to improve classifier performance.[24] It should be noted that not all models are capable of handling both discretized and continuous variables, and thus both methods were not applied to all modeling methods. Additionally, some methods, such as certain kinds of neural networks or decision trees, may dynamically convert numeric variables into binary or categorical variables as part of their modeling process.

Multiple models were constructed on the dataset to determine optimal performance using both native, built-in KNIME models as well as models incorporated from WEKA (Waikato Environment for Knowledge Analysis; Version 3.5.6).[20] Models were generally run using default parameters, though some experimentation was performed. Models tested included Naïve Bayes,[20] HNB (Hidden Naïve Bayes),[25] AODE (Aggregating One-Dependence Estimators),[26] Bayesian Networks,[20] Multi-layer Perceptron neural networks,[20] Random Forests,[27] J48 Decision Trees (a variant of the classic C4.5 algorithm),[28] Log Regression, and K-Nearest Neighbors.[20] Additionally, ensembles were built using a combination of Naïve Bayes, Multi-layer Perceptron neural network, Random Forests, K-nearest neighbors, and logistic regression, employing forward selection optimized by AUC (area under the curve).[29] Voting by committee was also performed with those same five methods, based on maximum probability.[30] Voting by committee is a meta-modeling technique (like ensemble) that combines multiple models by allowing them to "vote" for the winning classification. It seeks to take advantage of the strengths of different modeling approaches while minimizing their drawbacks. Due to the number of models used, detailed explanations of individual methods are not provided here for brevity, but can be found elsewhere.[20,23]

The last step was evaluating model performance to rule out the possibility that statistical findings may be an artifact of capitalization on chance, which was performed using 10-fold cross-validation.[20] All models were evaluated using multiple performance metrics, including raw predictive accuracy; variables related to standard ROC (receiver operating characteristic) analysis, the AUC (area under the curve), the true positive rate, and the false positive rate[31] and Hand's H.[32] For readers unfamiliar with ROC analysis, it is a form of signal analysis where AUC represents the area under the curve of a plot of true positives and false positives (we refer interested readers to the excellent introduction on the subject by Fawcett 2004).[31] The data mining methodology and reporting is in keeping with recommended guidelines,[33] such as the proper construction of cross-validation, incorporation of feature selection within cross-validation folds, testing of multiple methods, and reporting of multiple metrics of performance, among others.

Additionally, some better performing models were evaluated using feature selection prior to modeling (but within each cross-validation fold). Feature selection is a key component in filtering out noisy and/or redundant variables from datasets and building parsimonious, explanatory models that retain generalizability. Various methods were attempted: univariate filter methods (Chi-squared, Relief-F), multivariate subset methods (Consistency-Based –Best First Search, Symmetrical Uncertainty Correlation-Based Subset Evaluator) and wrapper-based (Rank Search employing Chi-squared and Gain Ratio). The advantages and disadvantages of these different types of feature selection are well-addressed elsewhere.[34]

## **Results**

The results of the various combinations of modeling method and discretization can be seen in Table 4, sorted by AUC. The highest accuracies are between 70-72%, with AUC values ranging between .75-.79. It should also be noted that the Spearman's rank-order correlation between AUC and Hand's H was .977 (p<.01), indicating little divergence between the two measures, at least on this particular dataset. Hand[32] has indicated that these two measures will diverge when misclassification costs vary across methods. There was no evidence of that in this case, or at least none indicating that the divergence was significant. These initial results suggest a predictive capacity of the current EHR data within Centerstone. It is suspected that utilizing more sensitive outcome measures designed to specifically measure change over time will improve this capacity.

**Table 4: Model Performance**

| 10X Cross-Val (partitioned) | | | | | | |
|---|---|---|---|---|---|---|
| **Model** | **Binning** | **Accuracy** | **AUC** | **TP rate** | **FP rate** | **H** |
| AODE | CAIM | 72.3% | 0.777 | 74.6% | 32.6% | 0.274 |
| Lazy Bayesian Rules | CAIM | 71.2% | 0.774 | 75.2% | 36.2% | 0.270 |
| Naïve Bayes | CAIM | 71.6% | 0.771 | 76.5% | 36.5% | 0.271 |
| Bayes Net - K2 | CAIM | 70.7% | 0.769 | 75.4% | 37.4% | 0.255 |
| Bayes Net - K2 | Bin Target | 70.4% | 0.768 | 75.7% | 38.1% | 0.256 |
| Ensemble | CAIM | 70.9% | 0.760 | 76.9% | 38.1% | 0.245 |
| Naïve Bayes | Bin Target | 68.6% | 0.759 | 74.7% | 41.0% | 0.241 |
| Bayes Net - TAN | CAIM | 70.0% | 0.757 | 73.3% | 37.0% | 0.230 |
| Bayes Net - TAN | Bin Target | 69.7% | 0.756 | 73.4% | 37.6% | 0.239 |
| MP Neural Net | CAIM | 70.7% | 0.753 | 75.6% | 37.6% | 0.227 |
| Ensemble | Bin Target | 70.2% | 0.750 | 74.5% | 37.6% | 0.220 |
| Classif via Linear Reg | Bin Target | 68.8% | 0.749 | 71.5% | 37.6% | 0.236 |
| MP Neural Net | Bin Target | 69.5% | 0.747 | 73.0% | 37.7% | 0.237 |
| K-Nearest Neighbor | CAIM | 69.5% | 0.738 | 73.6% | 38.4% | 0.209 |
| Vote | CAIM | 68.1% | 0.736 | 72.7% | 40.5% | 0.201 |
| Random Forest | Bin Target | 66.0% | 0.724 | 70.3% | 43.1% | 0.204 |
| Random Forest | CAIM | 67.8% | 0.722 | 71.7% | 40.1% | 0.190 |
| Log Regression | CAIM | 67.8% | 0.721 | 77.7% | 47.9% | 0.181 |
| Log Regression | Bin Target | 67.1% | 0.712 | 71.7% | 41.7% | 0.180 |
| J48 Tree | CAIM | 68.1% | 0.681 | 71.5% | 39.4% | 0.169 |
| Vote | Bin Target | 63.4% | 0.661 | 76.2% | 57.1% | 0.124 |
| J48 Tree | Bin Target | 66.9% | 0.654 | 72.4% | 32.6% | 0.149 |
| K-Nearest Neighbor | Bin Target | 63.8% | 0.636 | 65.9% | 44.2% | 0.079 |

These models were then applied to a series of pre-determined "service packages" that most typical patients receive. The results of one of the higher performing models by AUC (Bayesian Network – K2) were used to generate predictive information for the clinician at the time of intake. Implementation with the live system is being addressed in a separate, upcoming study (data not shown). However, examples of these predictions (based on actual data) can be seen in Figures 2 and 3.

**Figure 2: Example 1 of treatment recommendations using pre-set "service packages"**

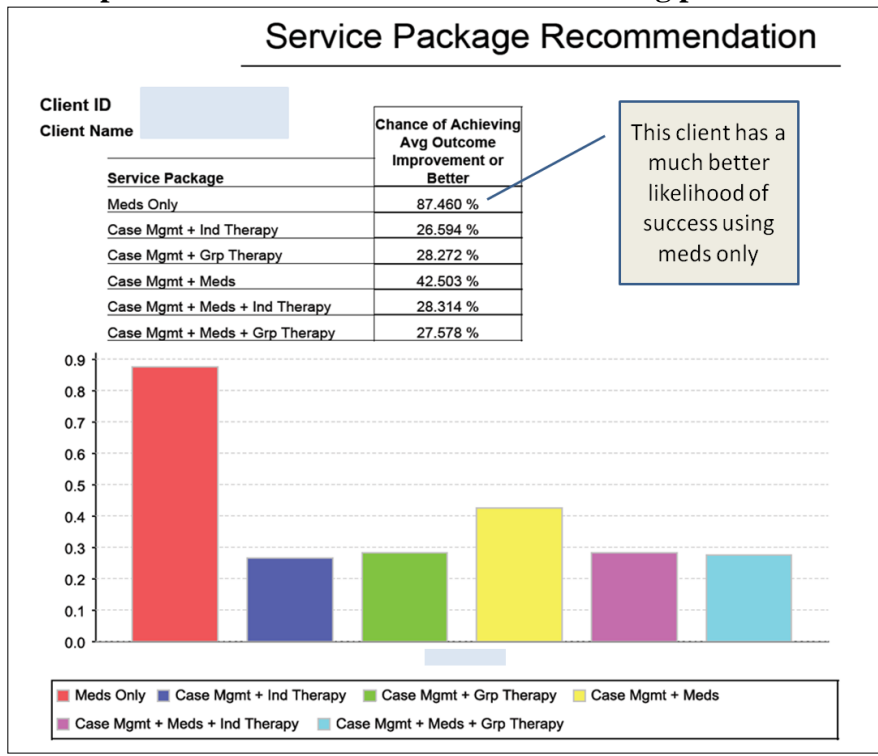

**Figure 3: Example 2 of treatment recommendations using pre-set "service packages"**

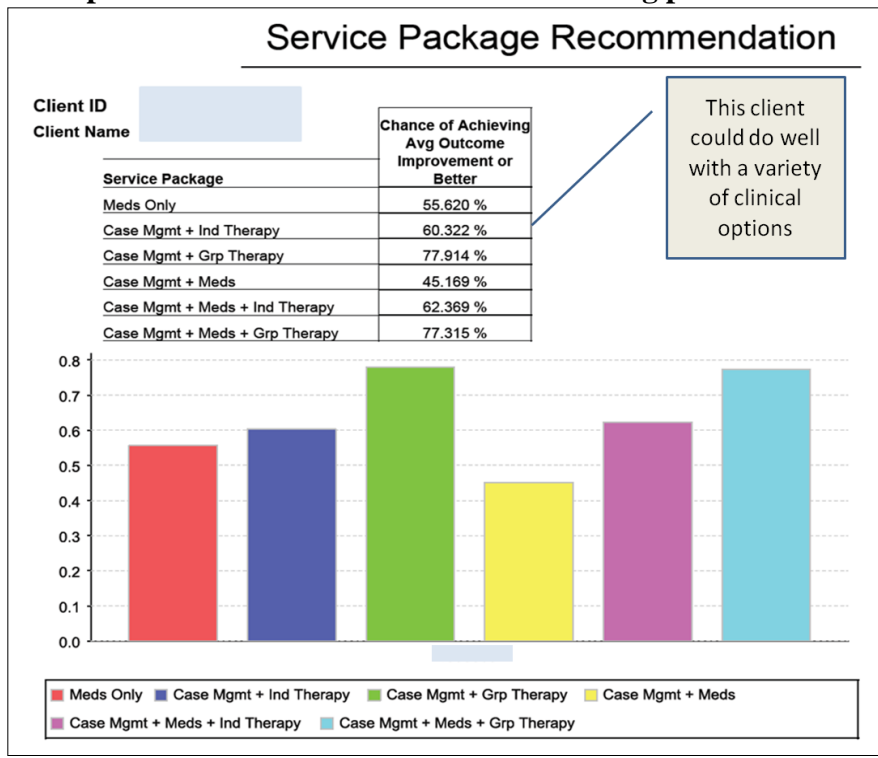

The results of feature selection were mixed (data not shown).  Although some methods were able to produce similar performance using smaller, more parsimonious feature sets than the full feature set models (most notably wrapper-based approaches), they generally did not improve performance significantly.  Additionally, the selected feature sets displayed a marked degree of variability across methodologies.  This is a common issue, to be expected with complex problems.[35] In many domains there are potentially multiple models/feature sets that can produce comparably good results.  In terms of individual predictors (as assessed by the chi-squared feature selection method), baseline CARLA score was the most significant (Odds Ratio: 4.1, 95% Confidence Interval: 1.83-6.37).  In other words, patients with lower baseline scores were 4.1 times more likely to experience above-average outcome improvement than those patients with higher baseline scores.   Other variables with consistently significant impact included payer (possibly as a proxy for socio-economic status), diagnostic category, county/location, and provision of case management services.  However, given the mixed improvement using feature selection and small sample size, individual predictor variable performance should be taken with caution at this juncture.

**<u>Conclusions</u>**

Recent years have the seen the proliferation of electronic health records (EHRs) across the mental healthcare field and the healthcare industry in general.  The current challenge is turning data collected within EHRs into information useful for healthcare providers and addressing the 17-year research-practice gap in healthcare.  Technology utilizing predictive modeling and data mining has the potential to transform data about the past into information about the future to improve clinical care.  Indeed, ongoing research in a large community-based mental health provider has produced models that are accurate over 70% of the time (i.e., for 7 out of 10 patients), even without enhancing data collected in the EHR specifically for predictive purposes.  Through such an approach the potential of EHRs can be realized.  Long term, this will significantly impact the way research is incorporated into practice.

Predictive modeling can produce tailored recommendations that adapt to variation of real-world populations, and even individuals.   New innovations and recommendations for individualized care can literally be integrated into predictive models overnight, as opposed to the decades that research evidence often takes to diffuse into common practice.   Without individualized care recommendations that have the capacity to rapidly incorporate changing evidence, adoption of evidence-based practice and treatment guidelines will likely continue to lag. While there are many barriers to adoption of systematic treatment recommendations, one of the primary failings of common treatment recommendations is that they are based on statistical averages (e.g., "70% of people improve with medication X").[5] Practicing clinicians are keenly aware that treatments that have been shown to be highly effective in clinical trial and research populations are not always directly transferable to individuals in real-world settings. Conversely, data mining models actually benefit from natural variation in clinical practice. This is a major difference between evidence-based models and practice-based evidence, at least as they are currently conceived and implemented. Ultimately, it is likely that greater utilization of evidence-based practices will depend on the incorporation of systematic adaptations based on practice-based evidence. As LW Green noted, "what practitioners in clinical, community, and policy-making roles crave, it appears, is more evidence from practices or populations like their own, more evidence based in real time, real jurisdictions, typical patients, without all the screening and control …"[3]

The quickest way to disseminate research findings into practice is to build a framework that allows research to be conducted around ongoing clinical practice, without interfering with day-to-day clinical workflow. The research can then be utilized to provide decision support and feedback functionality to clinicians. We term this "adaptive" clinical decision support. The approach here is moving toward such a framework. However, it is fraught with many challenges. First, while this paper emphasizes the adaptability of the system, one important consideration is the embedding of certain clinical standards around treatment. In essence, this inserts hard-coded rules into the system, holding constant those treatment variables while varying other treatment recommendations around them. Indeed, experience at Centerstone has underscored how critical this ability may be to actual clinician adoption. The system also needs to provide information to help support provider decision-making process, providing the right information at the right time in the right context. The principles of mental workload and situational awareness (derived from human factors research) are critical in understanding how clinicians may interact with such information. An engaging user interface that can incorporate clinical standards, provide appropriate information, and accommodate known human factor issues is key.[36]

This approach focuses on creating an environment where a clinician's natural course of adaptation is toward data-driven models. In a sense, it relies heavily on natural evolutionary principles and less on artificial constructs of directly altering clinical behavior, the latter of which have been shown to have limited sustainability.[6] While this may be a somewhat atypical approach to implementation, it parallels many current models of behavioral change in psychotherapy, decision support technology implementation, and artificial intelligence.[36-38] Utilizing an evolutionary construct invites application of a wealth of evolutionary models and mathematical constructs that, once adapted, may provide an analytical foundation for understanding the process of implementation in general.

The greatest challenge to this approach is building meaningful models for clinicians that answer relevant questions.  However, it also represents a significant opportunity to delve into more specific questions.  In the work described here, the initial model was constructed across all diagnoses - including diagnosis used as a predictor variable – but work is proceeding toward construction of models that make personalized clinical predictions within diagnostic-specific groups.  Furthermore, we are analyzing more finite considerations of clinical practice, moving from – "does the patient need medications?" – to "which medications are most likely to be effective for this particular individual?"  Mixing genetic (e.g. microarray) and clinical indicators, rather than using one or the other, is the most likely long-term avenue, although if and how these data sources should be combined is still a subject of intense debate.[18,19]

One limitation of this approach is that it requires large populations, diversity in clinical practice, and reliable data.  A small medical practice or group practice might not be able to generate enough data to produce reliable and replicable findings. It is therefore important for small provider organizations to consider how to aggregate their data so that predictive models may be developed and fed back into local EHRs (for instance, via Health Information Exchanges [HIEs]).[39] Privacy and security of health information will be paramount, so as to strike the optimal balance between protecting individual privacy and the collective benefit of data aggregation toward potential, meaningful advances in care.

The case study presented here demonstrated the feasibility of building clinically predictive models using data already existent in the EHR.  New studies are currently underway evaluating these approaches with patient-reported outcomes and for data-driven decision support in controlled pilot settings for patients with depression.[40] We are also developing a national data warehouse to share EHR data across several major mental healthcare providers from multiple states, in partnership with Centerstone Research Institute's Knowledge Network, a technology-based alliance of providers, academic researchers, and industry leaders.   Funding is also being sought to develop a gene expression database on a large portion of Centerstone's clinical population, likely starting with patients with depressive disorders or schizophrenia.  These efforts aim to improve and validate the approaches laid out herein.

Implementation of EHRs is only the first step in using technology to advance health care, though that first step has proven to be challenging.  Indeed, even the popular media is picking up on this fact ("Little Benefit Seen, So Far, in Electronic Patient Records" *New York Times*, 11/15/2009).  Without actually modeling the data, an EHR is only informative of what happened in the past, not predictive of what might happen in the future.  Without that predictive capacity, it cannot be used by clinicians as actionable information.  In that sense, without predictive modeling/data mining, EHRs are essentially just copies of paper-based records stored in electronic form.  There is limited expected gain in terms of clinical outcomes, quality, and efficiency.  Predictive modeling/data mining turns an EHR into the decision support tool as was envisioned in the beginning, opening doors to possibilities for technologies to be built on top of the EHR that can enhance clinical care and improve efficiency.  These technologies include behind-the-scenes data infrastructures, warehouses, and potentially national repositories[11] that ultimately merge research and practice into one cohesive process.

**Acknowledgements**

This commentary and related research is funded by the Ayers Foundation and the Joe C. Davis Foundation. The opinions expressed herein do not necessarily reflect the views of Centerstone or its affiliates. The authors have no conflict of interest with the subjects described herein.